\PassOptionsToPackage{table}{xcolor}
\documentclass[journal]{IEEEtran}

\IEEEoverridecommandlockouts

\usepackage{graphicx}
\usepackage{epsfig}
\usepackage{mathptmx}
\usepackage{times}
\usepackage{amsmath}
\usepackage{amssymb}
\usepackage{algorithm}
\usepackage{algpseudocode}
\usepackage{tabularx}
\usepackage{booktabs}
\usepackage{siunitx}
\usepackage{threeparttable}
\usepackage{textcomp}
\let\labelindent\relax
\usepackage{enumitem}
\usepackage{caption}
\usepackage{subcaption}
\usepackage{url}
\usepackage{xcolor}
\usepackage{pifont}
\usepackage{tikz}
\usepackage{microtype}
\usepackage{adjustbox}
\usepackage{multirow}
\usepackage[hidelinks]{hyperref}
\usepackage{etoolbox}
\patchcmd{\abstract}{---}{: }{}{}
\patchcmd{\IEEEkeywords}{---}{: }{}{}

\definecolor{hddUgray}{gray}{0.92} 
\definecolor{totalscolor}{RGB}{225,236,248}
\definecolor{unmakeblue}{RGB}{225,236,248}

\newcommand{\newadd}[1]{{\color{black}#1}}

\newcommand{\adipchange}[1]{{\color{black}#1}}
\newcommand{\adipchangetwo}[1]{{\color{black}#1}}
\makeatletter
\let\adipchangeoriginal\adipchange
\DeclareRobustCommand{\adipchange}{\@ifnextchar2{\adipchangetwodispatch}{\adipchangeoriginal}}
\def\adipchangetwodispatch2#1{\adipchangetwo{#1}}
\makeatother
\definecolor{unmakepink}{RGB}{205,36,117}
\definecolor{unmakegray}{RGB}{64,64,64}
\newcommand{\unmake}[1]{{\color{unmakegray}#1}}

\makeatletter

\newcommand{\hddT}{HDD}
\newcommand{\hddU}{HDD$[{\mathrm{U}}]$}

\title{\LARGE \bf
UNMAKE: Language-Agent Planning over Structured World States for Adaptive Dual-arm Robotic Disassembly
}

\author{%
Adip~Ranjan~Das, David~Neilan, and Maria~Koskinopoulou%
\thanks{%
Adip Ranjan Das, David Neilan, and Maria Koskinopoulou
are with the School of Engineering and Physical Sciences,
Heriot-Watt University, Edinburgh, UK
(e-mail: \texttt{ard2000@hw.ac.uk};
\texttt{ddn4000@hw.ac.uk};
\texttt{M.Koskinopoulou@hw.ac.uk}).%
}%
\\[0.6ex]
{\small
\textbf{Project website:}
\url{https://arm-2-lab.github.io/language_agent_disassembly_project/}%
}%
}

\begin{document}

\maketitle
\thispagestyle{empty}
\pagestyle{empty}

\begin{abstract}
Autonomous robotic disassembly supports scalable electronic-waste recycling, but fixed-sequence execution can fail when physical outcomes depart from the intended task progression. This paper presents \unmake{UNMAKE}, an adaptive dual-arm disassembly framework in which a language agent plans over a perception-grounded structured world state. Three perception modules estimate the scene configuration, refine fastener localisation, and verify the outcome of every skill invocation. The resulting state is the only dynamic input to a ReAct-based agent that selects one of three programmed skills and its target, without a supplied action sequence, removal-precedence instructions, or device identity. The framework is implemented on a dual-arm cell in which a manipulation arm stabilises and extracts components and a tooling arm performs precision unscrewing. In 45 trials on a hard disk drive (HDD), a mini~PC, and a laptop under nominal and altered starting conditions, \unmake{UNMAKE} increased device-averaged autonomous target-step completion from $58.1\%$ for a scripted fixed-sequence baseline to $89.7\%$ and reduced cascade failures by $90.0\%$. Ten further trials on an additional unknown HDD, not used for training, achieved $90.9\%$ autonomous completion.
\end{abstract}

\begin{IEEEkeywords}
Disassembly, 
Agentic Planning, LLMs,
Dual-Arm Manipulation, 
Perception for Manipulation, Sustainability
\end{IEEEkeywords}

\section{Introduction}
\label{sec:intro}
\IEEEPARstart{E}{lectronic} waste recovery requires scalable disassembly. Robotic automation can address this need, but end-of-life (EoL) devices pose particular difficulties: they are often worn, damaged, or inconsistent, even within one product line \cite{das2024towards}. Recent surveys identify product variability as a continuing challenge for robotic disassembly \cite{ameur2025review, daneshmand2023industry}. Unlike assembly under controlled component and fixture conditions, disassembly must contend with seized fasteners, missing components, and prior modifications, which can differ even between units of the same model \cite{poschmann2020disassembly}.

\begin{figure}[t!]
  \centering
  \includegraphics[width=0.9\columnwidth]{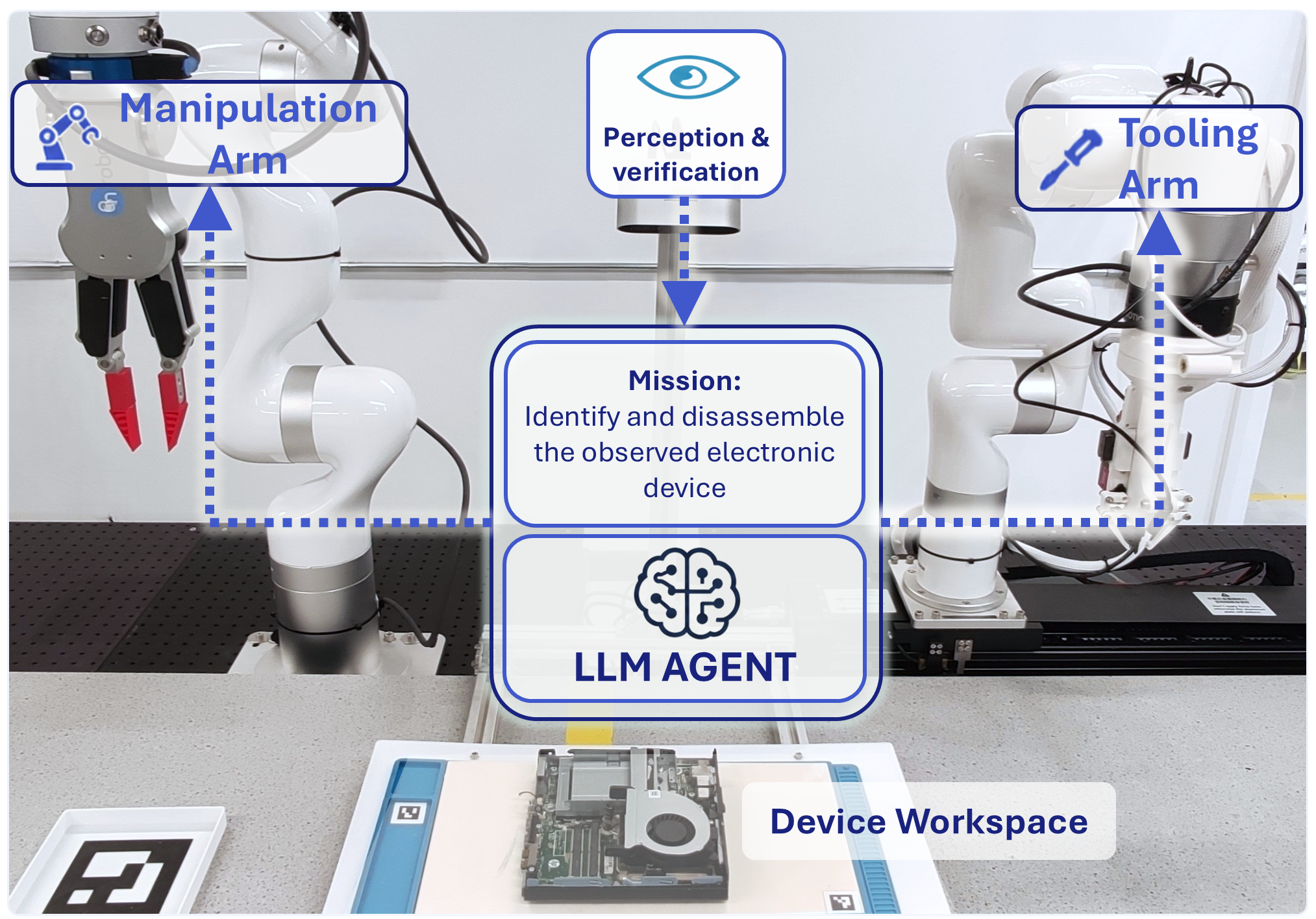}
  \caption{Experimental dual-arm disassembly cell. The manipulation arm handles
  stabilisation, reorientation, and extraction, while the tooling arm performs
  precision unscrewing under overhead visual perception.}
  \label{fig:expsetup}
\end{figure}

A fixed-sequence controller executes a predefined action order or a schedule computed before execution. Its main limitation appears once an operation fails: a screw that remains seated can block a later extraction if the controller proceeds as though removal succeeded. Part failures have accordingly been considered in disassembly sequence planning \cite{cui2023robotic}. Adaptive planners that verify outcomes and replan have also been reported ~\cite{vongbunyong2013cirp, cashmore2015rosplan, das2026egrap}, but they typically encode task knowledge in a hand-built, product-specific model. Dual-arm cells are well suited to disassembly, since one arm can stabilise the device while the other operates on it. Dual-arm studies include assembly-oriented task-sequence planning \cite{wang2022assembly} and teleoperated disassembly \cite{hathaway2023towards}; the present study instead addresses autonomous task-level skill selection from verified observations while retaining role-specialised arm control (Fig.~\ref{fig:expsetup}).

Large language models (LLMs) offer an alternative, as they can select robot skills without a hand-built symbolic task model~\cite{saycan2022arxiv}. In disassembly, however, LLMs have mainly been used to generate complete sequences before execution~\cite{erdogan2026intent}; grounding a language agent in verified, fastener-level outcomes on a physical dual-arm cell has received limited attention.

\begin{figure*}[h!]
  \centering
  \includegraphics[width=0.90\textwidth]{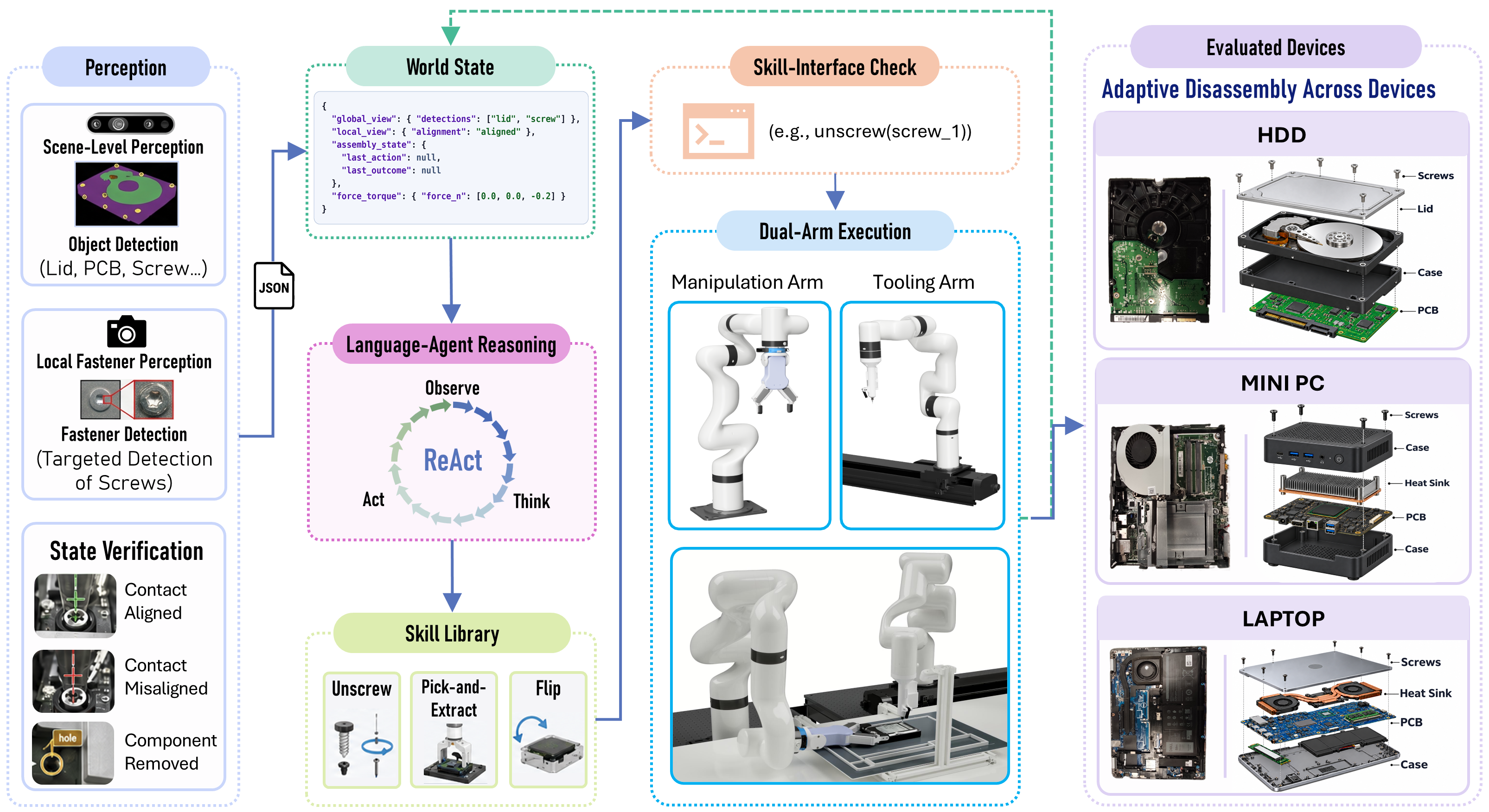}
  \caption{\unmake{UNMAKE} architecture for adaptive dual-arm disassembly. Perception
  updates the structured world state, which the planner uses to dispatch
  parameterised skills to the role-specialised execution layer.}
  \label{fig:concept}
\end{figure*}

This paper addresses online task ordering and recovery in physical dual-arm disassembly. The agent receives the device state, the mission, and the skill descriptions, but no ordered sequence, no explicit removal-precedence instructions (e.g., to remove screws before the lid or printed circuit board (PCB)), and no device identity. A structured world state, maintained through scene-level perception, local fastener perception, and post-action verification, grounds the planner's decisions; its fields are written by the perception and verification models rather than by the language model. Following a ReAct-style reasoning process~\cite{yao2022react}, the planner, an LLM-based language agent (``LLM agent'' in the figures), uses this state to select and parameterise reusable manipulation skills, so that the sequence adapts to the physical state of the device (Fig.~\ref{fig:concept}).

The main contributions of this work are as follows:

\begin{itemize}
    \item [-] A structured world state that serves as the only dynamic interface between perception and planning, combining scene-level components, target-fastener measurements, force-torque measurements, and verified execution outcomes in one planner-facing record.

    \item [-] A language-agent planner that constructs and revises the disassembly sequence online by selecting and parameterising task-level skills, checked against a predefined skill interface, according to the current world state, without a supplied device-specific sequence or explicit task-precedence instructions. Low-level manipulation skills remain conventionally programmed.

    \item [-] An execution loop in which every skill call is followed by outcome verification, and in which initial target failures, planner-selected recovery attempts, and physical assistance are recorded separately.

    \item [-] A physical evaluation on three structurally distinct devices (a hard disk drive (HDD), a mini~PC, and a laptop), using one perception stack trained on these units and one skill library, compared against a scripted vision-guided baseline on the same hardware, and supplemented by ten agentic trials on a different-brand HDD excluded from global-camera detector training. The comparison evaluates the combined execution-aware system through target-step completion, recovery, cascade failures, intervention, and cycle time; the contributions of individual components are not isolated.
\end{itemize}

The remainder of this paper is organised as follows. Section~\ref{sec:related} reviews related work. Section~\ref{sec:method} presents the proposed framework. Section~\ref{sec:expmeth} describes the experimental methodology, Section~\ref{sec:results} reports the results, and Section~\ref{sec:conclusion} concludes the paper.

\section{Related Work}
\label{sec:related}

\textbf{Adaptive Disassembly Planning.}
\label{sec:rw-adaptive}
Model-based planning represents task structure explicitly, for example through subassembly decomposition \cite{watson2019assembly}. Failure-aware disassembly planning further considers component accessibility and damage \cite{cui2023robotic}. A predefined sequence should be distinguished from a planner that recomputes its decisions after execution feedback. Predicate-based planning and feedback-driven policies can both support such adaptation; the relevant distinction here is how task ordering is obtained, not whether a closed loop is possible.

For example, eGRAP maintains a dependency graph during coordinated dual-arm HDD disassembly and updates it as the task proceeds \cite{das2026egrap}. \unmake{UNMAKE} targets the same need for execution-aware disassembly, but uses a language agent in place of an explicit graph scheduler and reasons over a single structured state holding current observations and verified action outcomes.

Other adaptive approaches monitor execution and revise decisions online: a cognitive agent that checks each state transition with vision and revises its plan~\cite{vongbunyong2013cirp, vongbunyong2016monitoring}, probabilistic planning that maintains a belief over hidden end-of-life conditions~\cite{baumgartner2025pomdp}, and reinforcement-learning policies that select the next operation under structural uncertainty~\cite{zhao2021rldsp}. These systems already make state-dependent decisions, but encode them in product-specific rules, models derived from computer-aided design (CAD) data, or policies trained per product.

\vspace{2pt}

\textbf{Perception for Disassembly.}
\label{sec:rw-perception}
Perception underpins autonomous disassembly. Prior work has combined vision and force sensing to localise fasteners on reflective metal \cite{diaz2022robotic}. Convolutional networks have since been trained to detect and classify screws and components for HDD recycling \cite{yildiz2019cnn, yildiz2020visual}. Transformer detectors such as DETR \cite{carion2020detr} and its real-time variants \cite{rfdetr2024} provide alternative detection architectures, and RAISE \cite{liu2025raise} demonstrated a perception-driven pipeline for end-of-life phones. Target localisation and post-action outcome verification serve different roles: the former grounds a skill spatially, whereas the latter supplies evidence about whether the intended change occurred. Both inform the planner-facing state in the proposed framework.

\vspace{2pt}

\textbf{Execution Monitoring and Replanning.}
\label{sec:rw-symbolic}
In classical planning, the world state is a set of predicates and actions are operators with preconditions and effects, typically written in PDDL~\cite{fox2003pddl21}. Execution monitoring compares the predicted and perceived states and triggers replanning when they diverge, as in ROSPlan~\cite{cashmore2015rosplan}; task and motion planning extends this to continuous geometry~\cite{garrett2021tamp}. These frameworks perform the same observe--act--verify--update cycle as the present work and offer formal guarantees provided that the domain model is correct, but predicates, operators, and groundings must be written for every skill and product family. The present agent instead selects skills from a typed but open state without hand-written operator effects. This removes the need for an explicit domain model, at the cost of formal guarantees. Accordingly, the world-state update in the proposed framework is perception-based rather than a predictive transition model.

\vspace{2pt}

\textbf{Language Models as Task Planners.}
\label{sec:rw-llm}
LLMs support several planning interfaces. SayCan combines language scores with skill affordances~\cite{saycan2022arxiv}; Code as Policies generates robot-control programs~\cite{liang2023code}. ProgPrompt uses programmatic action/object descriptions and executable examples~\cite{singh2023progprompt}, whereas Inner Monologue incorporates success and scene feedback through language~\cite{huang2022inner}. These developments are reviewed in~\cite{wang2025llmrobotics}. SELF-VLA integrates vision-language-action policies, explicit skills, and a corrective policy for disassembly~\cite{selfvla2025}. In the present work, a text-based agent selects task-level skills from a component--fastener--outcome record on a dual-arm cell. The contribution is this specific interface and implementation, not the closed-loop concept itself.

Other work detects plan--execution misalignment with a vision-language model (VLM)~\cite{guo2023doremi} or uses a VLM to diagnose failed subtasks before replanning~\cite{skreta2024replan}, and Statler keeps an explicit world state that is itself written by an LLM~\cite{yoneda2024statler}. Since LLMs alone have been shown to be unreliable planners, the LLM-Modulo framework argues for pairing them with external verifiers~\cite{kambhampati2024llmmodulo}. The present framework follows this pattern, but its feedback arrives as typed state fields written by dedicated perception and verification models, rather than as free-form text, VLM answers, or LLM-predicted effects. This state is therefore a structured, perception-grounded semantic record, not a predicate-based symbolic state.

\vspace{2pt}

\textbf{Language Models in Disassembly.}
\label{sec:rw-fm-disassembly}
In disassembly, LLMs have mainly been used before execution, to evaluate sequences~\cite{xia2024mfglet} or to generate candidate removal sequences that are then checked by an LLM verifier~\cite{erdogan2026intent}. Closest to this work, Ude~\emph{et al.} use a VLM to select the next battery-removal action from a skill library based on the image, a symbolic state, and a check of the previous action~\cite{ude2025ram}. \unmake{UNMAKE} differs in that the planner-facing state is updated by dedicated detection and verification models rather than by the reasoning model inspecting images, that fastener-level outcomes are verified with a tool camera, and that one perception stack, state schema, and skill library are evaluated physically on three structurally distinct devices against a fixed-sequence baseline.

\section{Language-Agent Planning over Structured World States}
\label{sec:method}

The proposed framework formulates robotic disassembly as sequential decision-making over an evolving structured world state. As illustrated in Fig.~\ref{fig:concept}, this representation connects perception, language-agent reasoning, and dual-arm execution. Perception estimates the current device configuration and verifies the outcome of each skill invocation; the planner uses the resulting structured state, the mission, and the skill descriptions to select one executable skill; and the verified post-action observations update the state before the next decision. The disassembly sequence is therefore not specified in advance but is determined online from the observed device state, without a predefined task graph.

\begin{figure*}[!t]
  \centering
  \includegraphics[width=0.95\textwidth]{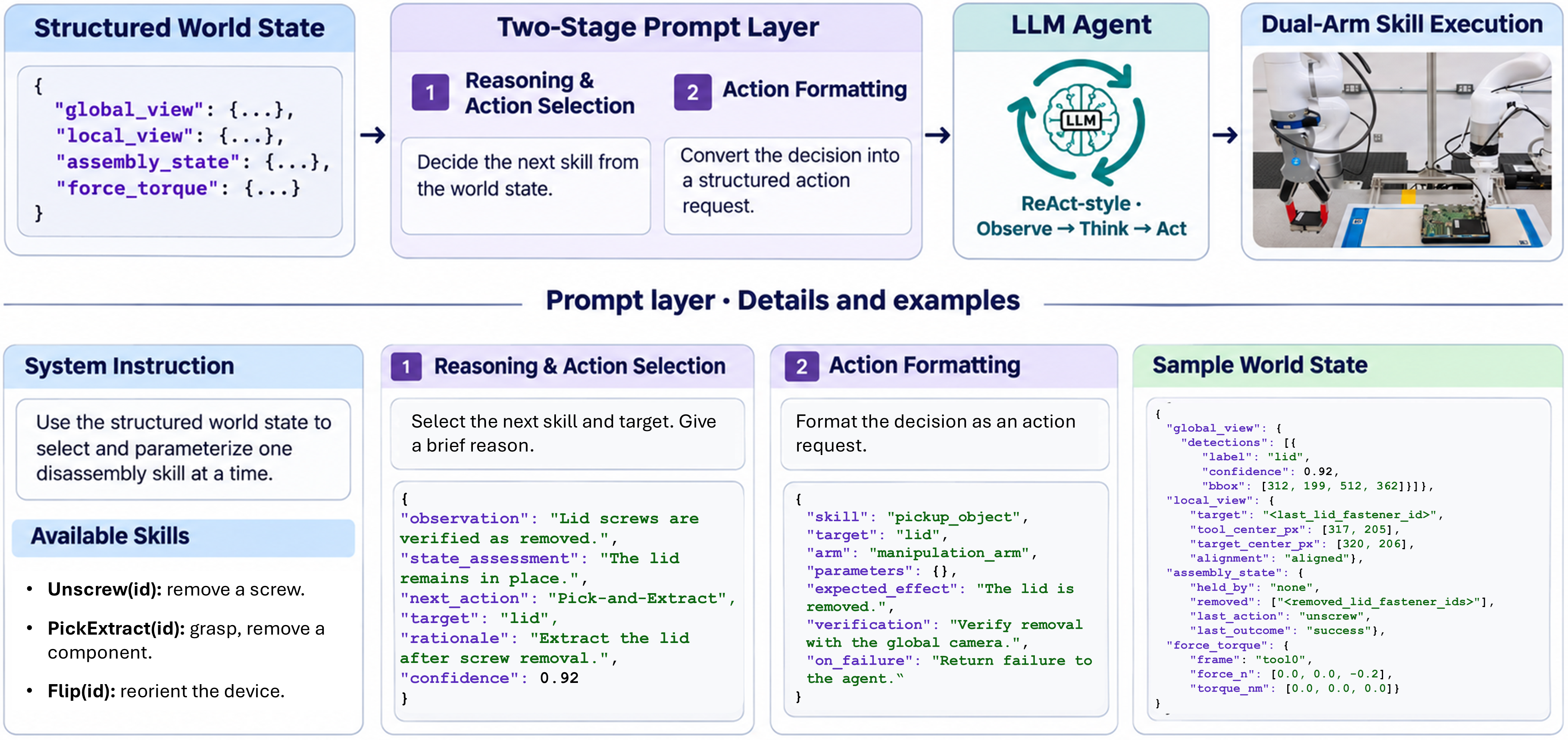}
  \caption{Structured world state and two-stage prompting interface for task-level robotic disassembly. The system instruction specifies the supported skills; the reasoning stage selects a skill and target from the world state, the formatting stage produces a parameterised request, and the request is checked against the skill interface before dispatch to the dual-arm execution layer. The prompts and JSON fields are condensed excerpts of runtime messages.}
  \label{fig:prompt_interface}
\end{figure*}

\subsection{Perception-Based World-State Construction}
\label{sec:worldstate}

The structured world state is the sole dynamic interface between perception and planning. It converts heterogeneous visual observations, geometric measurements, and execution outcomes into a unified structured semantic representation for the language agent, which receives these fields rather than raw images or free-form scene descriptions.

At each planning step $t$, the world state is represented as
\begin{equation}
S_t=\{O_t,F_t,E_t,W_t\},
\label{eq:world-state}
\end{equation}

where $O_t$ contains scene-level component information, $F_t$ contains information on the active target fastener, $E_t$ records verified execution outcomes and the current assembly state, and $W_t$ contains force--torque measurements. In the implementation, $S_t$ is serialised as a JSON record (Fig.~\ref{fig:prompt_interface}): \texttt{\small global\_view} ($O_t$) holds detection labels, confidence scores, and bounding boxes; \texttt{\small local\_view} ($F_t$) holds the active target, the tool and target centres, and alignment information; \texttt{\small assembly\_state} ($E_t$) records the currently grasped component, the removed components, and the latest action and its verified outcome; and \texttt{\small force\_torque} ($W_t$) holds force and torque in the sensor frame. All fields are written by the perception and verification modules described below, not by the language model.

\textit{\textbf{Scene-level perception}} populates $O_t$ with the visible device components and their class labels, confidence scores, and image bounding boxes. In addition to giving the planner a scene-level description of the current device configuration, it supplies coarse fastener locations that initialise subsequent manipulation, expressed in the common robot frame through the eye-to-hand calibration described in Section~\ref{sec:testbed}.


It uses RF-DETR-Large~\cite{rfdetr2024}, trained at $1088\times1088$ resolution on eleven classes (battery, case, core holder, HDD holder, heat sink, hole, lid, PCB, PCB holder, SSD holder, and screw) covering the components, fasteners, and fastener holes present across the three devices, and applied to images from the overhead global camera.

\textit{\textbf{Local fastener perception}} refines this information immediately before physical interaction. Rather than estimating the complete scene, it focuses on the active target and measures the image-plane offset between the screwdriver tip and the selected fastener. It uses RF-DETR-Small~\cite{rfdetr2024}, trained at $640\times640$ resolution on four classes (hole, screw, screw head, and tool head) and applied to images from the tool camera. The offset between the detected screw-head and tool-head centres populates $F_t$ and serves as the error signal for image-based visual servoing during precision unscrewing. A proportional-integral-derivative (PID) controller regulates this alignment error with a $\pm 5$-pixel tolerance and a $30$~s timeout.

Together, the two detectors form a two-level localisation process: the global view provides task context, and the tool view provides the precision required for tool--fastener alignment.

\textit{\textbf{State verification}} follows every skill invocation and evaluates whether the expected task-level transition has occurred before the planner commits to the next decision. Rather than assuming successful execution, the framework verifies the resulting change in the device and records the outcome in $E_t$. Fastener-level outcomes are assessed by a YOLOv11-Nano image classifier~\cite{yolo11_ultralytics}, applied to tool-camera images at $224\times224$ resolution and trained on three classes. It distinguishes three states: \emph{contact aligned}, in which the screwdriver tip is correctly seated on the screw head; \emph{contact misaligned}, in which the tip is offset from the screw head; and \emph{component removed}, in which the target fastener is no longer present. Component-level outcomes, such as an extraction or a reorientation, are verified from the scene-level detections of the global camera. Subsequent planning decisions are therefore conditioned on the verified device state rather than on the expected outcome.

\begin{figure}[t!]
  \centering
  \includegraphics[width=0.9\columnwidth]{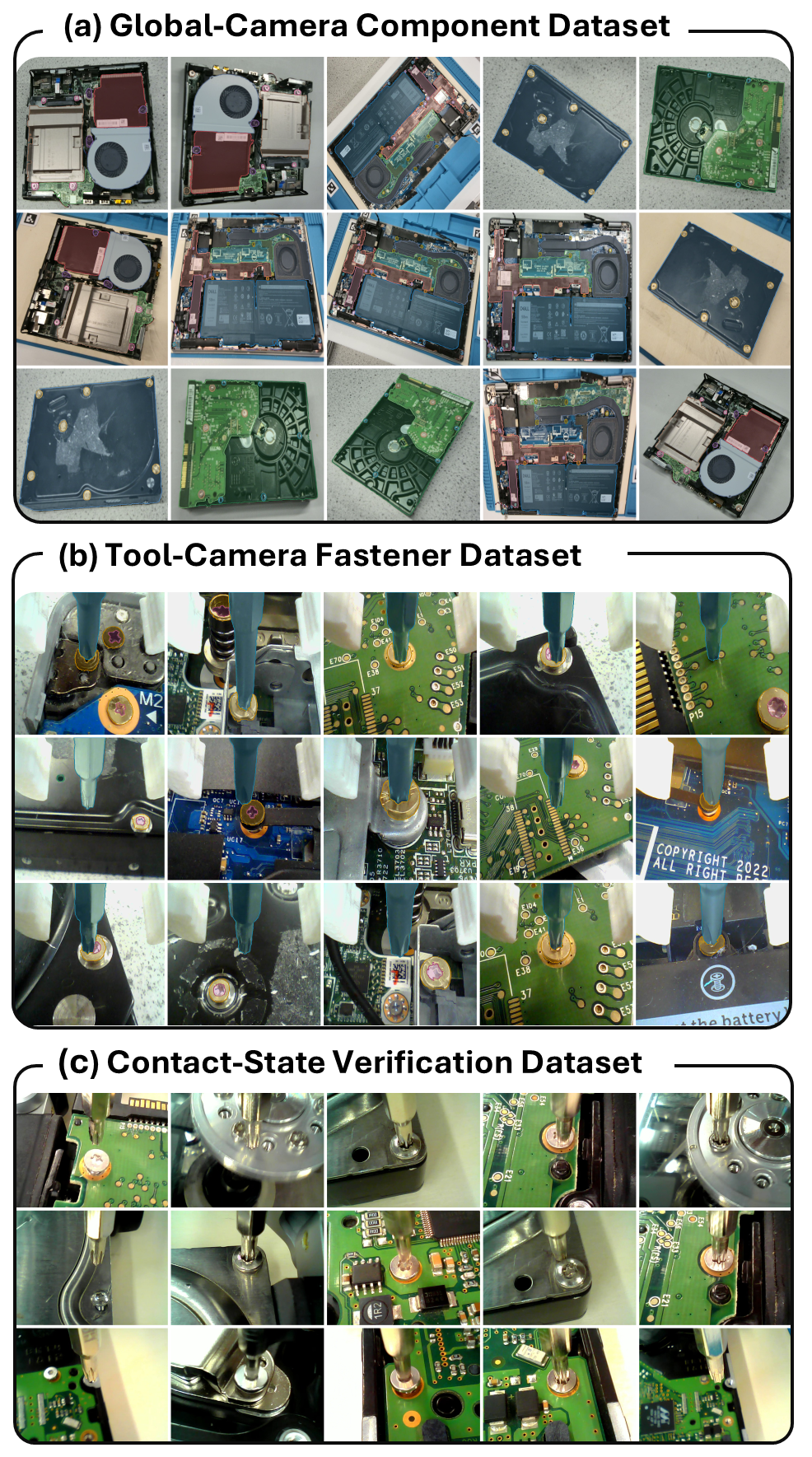}
  \caption{Representative samples from the three perception datasets: (a) global-camera views of the Western Digital HDD, HP mini~PC, and Dell laptop; (b) tool-camera views for fastener and tool-head localisation; and (c) tool-camera images for state verification.}
  \label{fig:sample_training}
\end{figure}

\textit{\textbf{Perception datasets and training.}} The three networks (i.e., RF-DETR-Large, RF-DETR-Small, and YOLOv11-Nano) are trained independently on images of three physical units covering three device categories and two manufacturers: one Western Digital HDD, one HP mini~PC, and one Dell laptop. Each unit contributed images to all three datasets and was also used in the disassembly experiments; the labels denote components, fasteners, and physical states, rather than brands. The scene-level and local fastener datasets each contain 800 raw images, and the state-verification dataset contains 175 raw images, all annotated in Roboflow~\cite{roboflow2024}. Each dataset was first split at the level of raw images into training, validation, and test sets (70/20/10), stratified by class, giving 560/160/80 images for each detection dataset and 122/35/18 for the state-verification dataset. Augmentation (rotation between $-15^\circ$ and $+15^\circ$, horizontal and vertical shear of $\pm2^\circ$, hue adjustment between $-15^\circ$ and $+15^\circ$, saturation adjustment of $\pm25\%$, brightness adjustment of $\pm20\%$, and mosaic augmentation) was then applied to the training set only, tripling it to 1,680 and 366 samples, respectively; the validation and test sets contain only unseen, unaugmented images. Model selection used the validation set, and each test set was evaluated once. The test images were therefore not seen during training, but were captured from the same three physical units; operation on a device not represented in the training data is evaluated separately on the \hddU{} (Section~\ref{sec:results-planning}). Test-set results are reported in Section~\ref{sec:results-perception}, and Fig.~\ref{fig:sample_training} shows representative samples.

\subsection{Language-Agent Reasoning}
\label{sec:agent-reasoning}

Given the current world state $S_t$, the planner selects the next task-level action
\begin{equation}
a_t=(k_t,\theta_t)=\pi(S_t),
\label{eq:planner-policy}
\end{equation}
%

where $\pi$ denotes the planning policy, $k_t$ is one skill from the predefined library, and $\theta_t$ denotes the identifier of the selected target.

The language agent selects the skill and target, while the target position is obtained from $S_t$ and low-level execution parameters are supplied by the programmed manipulation skills.
The corresponding planner-facing skill signatures are \texttt{Unscrew(id)}, \texttt{Pick-and-Extract(id)}, and \texttt{Flip(id)}.

The policy $\pi$ is implemented as a language agent in LangGraph~\cite{langgraph2024} with GPT-5.1~\cite{openai2025gpt51}, 
following the ReAct reasoning paradigm~\cite{yao2022react}, in which the model produces an explicit reasoning trace before committing to an action. The agent receives the mission (‘Identify and disassemble the observed electronic device’, Fig.~\ref{fig:expsetup}), the current state $S_t$, and descriptions of the three skills, but no ordered disassembly sequence, no explicit removal-precedence instructions, and no device identity. Because all perceptual information is encoded in structured fields within $S_t$, the model operates on task semantics rather than raw visual observations. Task ordering therefore relies on general knowledge acquired during pretraining, conditioned on the observed state.

 Each decision is produced in two prompt stages (Fig.~\ref{fig:prompt_interface}). In the reasoning stage, the model selects a supported skill and target and returns \texttt{\small observation}, \texttt{\small state\_assessment}, \texttt{\small next\_action}, \texttt{\small target}, \texttt{\small rationale}, and \texttt{\small confidence}. In the formatting stage, this decision is converted into a parameterised request specifying \texttt{\small skill}, \texttt{\small target}, \texttt{\small arm}, \texttt{\small parameters}, \texttt{\small expected\_effect}, \texttt{\small verification}, and \texttt{\small on\_failure}. These fields make the requested action and its expected outcome explicit at the skill interface.

The planner thus operates only at the task level: it selects and parameterises skills, while the programmed execution layer generates trajectories and performs low-level robot control. Every request is checked against the predefined skill interface before execution; this check verifies that the requested skill is registered in the skill library. Requests that fail this check are returned to the planner for revision; requests that satisfy it are dispatched to the execution layer. Invalid actions generate an error observation, allowing the agent to reconsider its next action.


\subsection{Adaptive Planning Through State Verification}
\label{sec:adaptive-sequencing}

Planning decisions follow from observed state transitions rather than from a fixed process graph. After action $a_t$ completes, the world state is updated according to
\begin{equation}
S_{t+1}=f(S_t,o_{t+1},a_t,y_{t+1}),
\label{eq:state-update}
\end{equation}

where $o_{t+1}$ denotes the observations acquired after execution and $y_{t+1}$ the outcome returned by the state-verification module (Section~\ref{sec:worldstate}). The update $f$ is deterministic and does not involve the language model: it refreshes $O$, $F$, and $W$ from the latest available perception and force--torque readings, and records $a_t$ and $y_{t+1}$ in $E$, updating the assembly state accordingly.

The updated state can confirm an expected outcome, reveal previously concealed structure, such as fasteners exposed by lid removal, or indicate that a target is still present. The planner applies the same decision policy~\eqref{eq:planner-policy} to $S_{t+1}$; no separate recovery mode is invoked.

For example, if a fastener remains seated after an \emph{Unscrew} invocation, the verification outcome is recorded in $E_{t+1}$, and the next decision is conditioned on it; the planner may reissue \emph{Unscrew} on the same fastener or select a different action. Structure exposed by a component removal enters the state through the same route. Task-level recovery is thus performed by the language agent, which issues a new skill invocation based on the verified outcome, rather than by device-specific recovery rules. The experimental protocol limits each target to three complete invocations (Section~\ref{sec:protocol}); within this budget, the choice of the next invocation is left to the planner. Brief alignment corrections remain within a single skill invocation and are not logged separately, whereas each subsequent planner-issued invocation is counted explicitly (Section~\ref{sec:protocol}).

\subsection{Dual-Arm Skill Execution}
\label{sec:skill-execution}

Actions selected by the planner are executed on the dual-arm platform shown in Fig.~\ref{fig:expsetup}. The manipulation arm handles stabilisation, device reorientation, and component extraction, while the tooling arm performs precision unscrewing.

The skill library comprises three skills. \emph{Unscrew} removes a threaded fastener by combining coarse localisation from $O_t$, visual servoing on $F_t$, force-guided engagement, and compliant extraction. The tooling arm drives the fastener while the screw-capture jaws at the screwdriver tip retain the loosened screw, so that it is removed together with the tool; when stabilisation is required, the manipulation arm secures the device as part of the same skill.
\emph{Pick-and-Extract} grasps and removes a component, and \emph{Flip} reorients the device to bring a concealed surface within reach; both use the manipulation arm alone. \emph{Pick-and-Extract} assumes that the component's fasteners have already been removed; because no precedence is supplied, this ordering must be inferred by the planner from $S_t$.

Each invocation is grounded in the target identities and spatial locations held in $S_t$, and each returns a status consumed by the update in~\eqref{eq:state-update}. A recorded invocation is one complete execution of a skill, including any internal alignment or contact corrections; first-invocation success therefore denotes completion within the initial invocation.

\section{Experimental Methodology}
\label{sec:expmeth}

This section describes the robotic platform, evaluation devices, comparative configurations, and trial protocol used to assess whether verified, state-informed replanning improves disassembly performance over fixed task sequencing.

\subsection{Robotic Testbed}
\label{sec:testbed}

\begin{figure}[t!]
    \centering
    \includegraphics[width=\columnwidth]{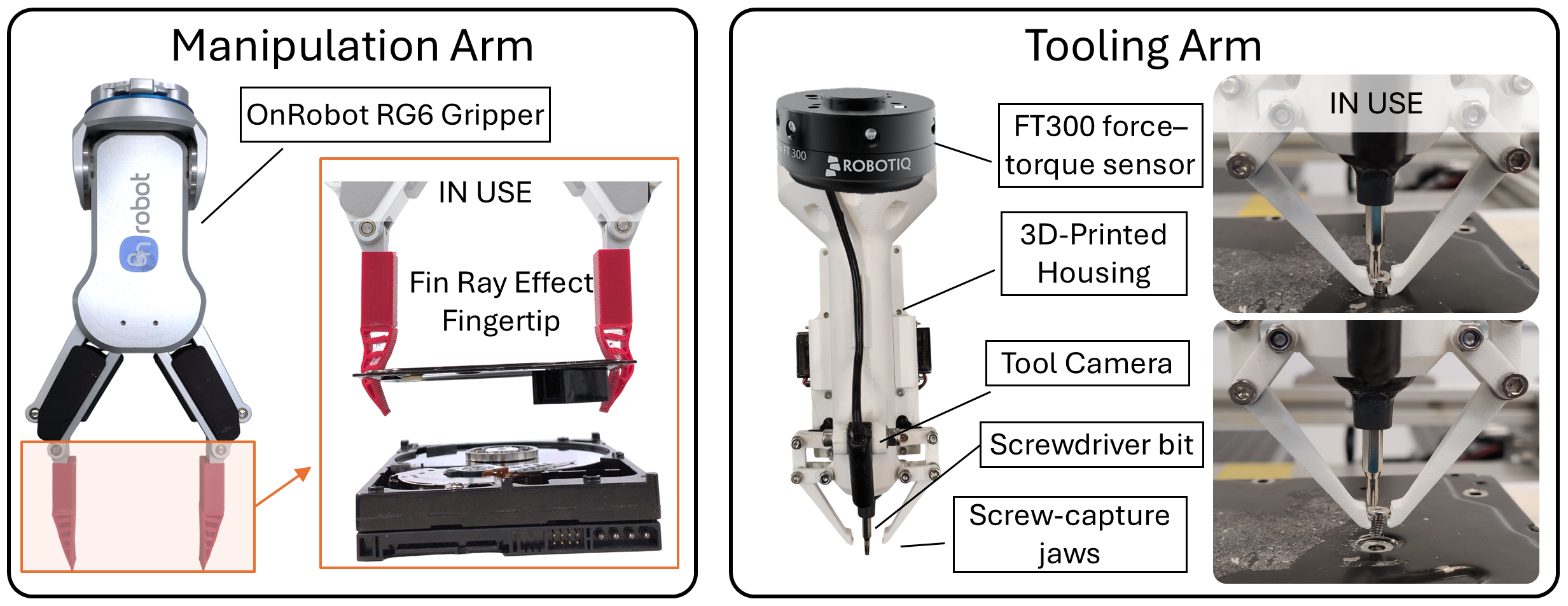}
    \caption{End-effectors of the dual-arm system: the compliant gripper used for component manipulation and the custom screwdriver used for fastener removal, with its tool camera, force--torque sensor, and screw-capture jaws.}
    \label{fig:grippers}
\end{figure}

Experiments are conducted on the dual-arm robotic disassembly cell shown in Fig.~\ref{fig:expsetup}. The two arms perform complementary roles. A 6-DoF UF850 manipulation arm stabilises, extracts, and reorients device components using an OnRobot RG6 parallel gripper, whose compliant Fin Ray-effect-inspired fingertips increase the contact area and accommodate geometric variation during grasping \cite{crooks2016finray,yang2021finray}. A 5-DoF xArm5 tooling arm performs fastener removal using a custom compliant electric screwdriver. The tool integrates a Robotiq FT300 force--torque sensor for contact monitoring, screw-capture jaws that retain loosened fasteners, and an endoscopic RGB camera (hereafter, the tool camera) for close-range localisation and visual servoing. The two end-effectors are shown in Fig.~\ref{fig:grippers}.

An overhead Orbbec Femto Bolt camera provides the global view used to initialise and update the scene-level component $O_t$ of the world state. The camera is registered to the common robot frame through eye-to-hand calibration: an ArUco marker mounted on the manipulation arm is observed at multiple robot poses, and the camera-to-robot transformation is estimated from these observations using the arm kinematics defined in the dual-arm URDF. Because both arms are described in the same URDF, global detections can be expressed in the shared workspace frame and used to position the tooling arm near a target, after which the tool camera performs fine alignment.

The complete system runs on ROS~2 Humble. EXOTica~\cite{exotica2019} generates collision-aware arm trajectories, while the perception modules send structured world-state updates to the task-level controller in JSON format.

\subsection{Test Devices and Disassembly Tasks}
\label{sec:test-devices}

The comparison between \unmake{UNMAKE} and the scripted baseline (Section~\ref{sec:protocol}) covers three screw-fastened electronic devices with distinct form factors, internal layouts, fastener types, and disassembly requirements; these comparison devices are the same physical units used to collect the perception datasets (Section~\ref{sec:worldstate}). Each fastener removal, component extraction, and reorientation constitutes one target step, executed by \emph{Unscrew}, \emph{Pick-and-Extract}, or \emph{Flip}, respectively:

\begin{itemize}[leftmargin=*, nosep]
    \item \textbf{HDD (\newadd{Western Digital-1T-3.5in, black}):} a compact 3.5-inch device in a rigid metal casing. An eleven-step task comprising six Torx T8 fastener removals, four component extractions (the lid, internal drive assembly, casing, and PCB), and one $180^\circ$ reorientation to expose the underside.

    \item \textbf{Mini~PC (\newadd{HP Elitedesk}):} a compact device with several internal compartments. Nine target steps: six Torx T5 fastener removals and extraction of the heat sink, HDD holder, and core holder. All targets are accessible from one face, and no reorientation is required.

    \item \textbf{Laptop (\newadd{Dell Latitude 5430}):} a wide, thin device with densely packed internals. Twelve target steps: eight Phillips fastener removals (nominal M2--M2.5) and extraction of the heat sink, SSD holder, battery, and PCB assembly. All accessible from the bottom face.
    
    \item \textbf{\hddU (\newadd{Seagate 1T-3.5in, silver}):} from a different manufacturer and not used in the training dataset, unseen during global-camera detector training, uses the same eleven-step task structure and is evaluated separately.
\end{itemize}

The three comparison tasks contain 32 target steps in total and collectively exercise all three skills.


\begin{figure*}[!t]
  \centering
  \includegraphics[width=0.9\textwidth]{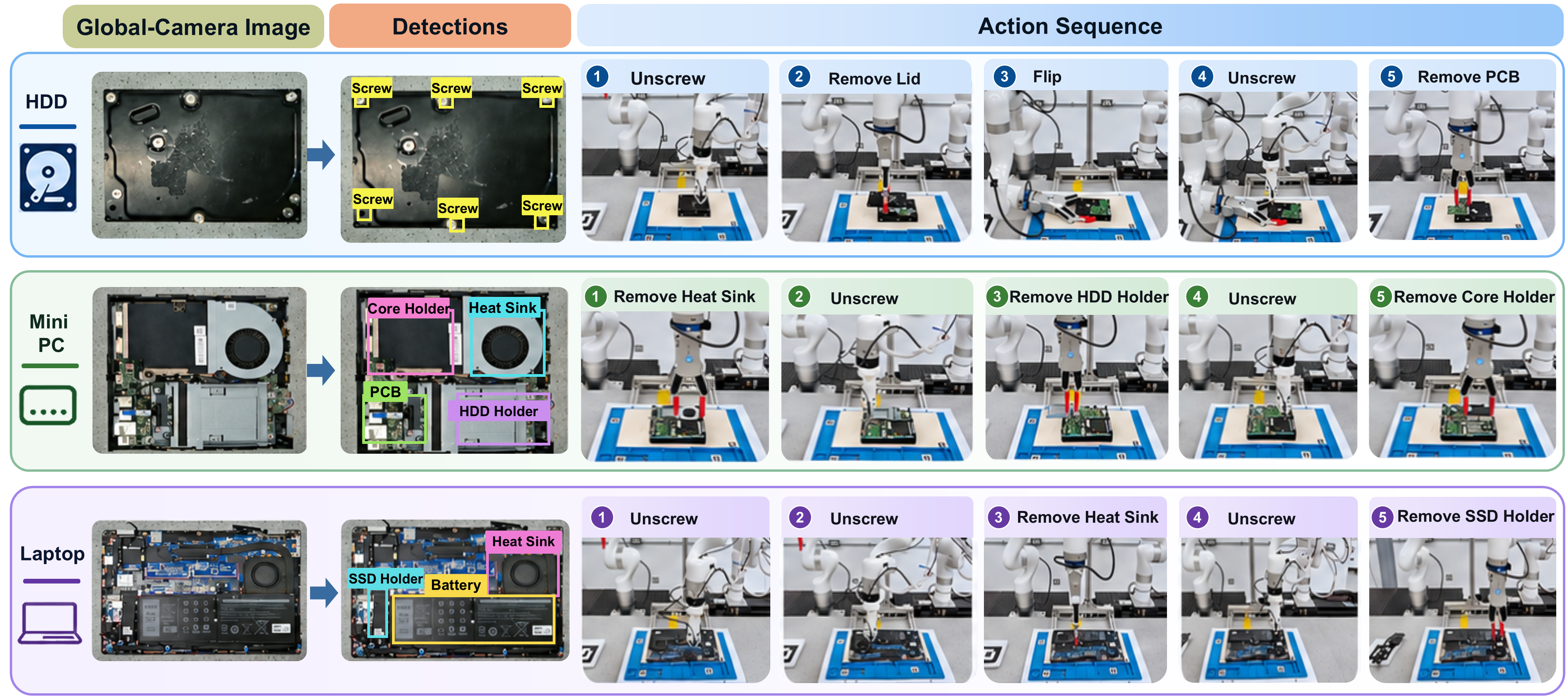}
  \caption{Representative perception-to-action examples for the HDD, mini~PC, and laptop. Each row presents a raw global-camera image, the corresponding detections, and selected skill executions. Labels beginning with ``Remove'' denote \emph{Pick-and-Extract}. The five displayed stages are excerpts rather than the complete nominal target-step sequence.}
  \label{fig:both_disassembly}
\end{figure*}

\subsection{Experimental Protocol and Evaluation Metrics}
\label{sec:protocol}

\unmake{UNMAKE} is evaluated on the four screw-fastened electronic devices, described above, with distinct form factors, internal layouts, fastener types, and disassembly requirements, and is compared on the same devices with a scripted fixed-sequence baseline representative of conventional fixed-sequence robotic disassembly~\cite{cui2023robotic}. Both systems share the same robotic platform, perception modules, motion planning, and manipulation skills; only the scripted baseline receives an ordered disassembly sequence. For each device, the authors constructed this sequence from manual disassembly and service documentation, following the nominal fastener--component precedence constraints: every fastener is removed before the component it secures. The baseline is vision-guided at the skill level, as target locations are obtained from scene-level and local fastener perception, but not at the task level: the sequence is not reordered in response to verified outcomes. After a failed invocation, the scripted baseline reissues the same skill on the same target within the three-invocation budget and marks the target step as failed if all three invocations are unsuccessful. The same sequence is used in all scripted trials, including altered starting conditions. The three-invocation failure rule also applies to targets that are already absent at the start of a scripted trial.

This baseline isolates the effect of task-level adaptation; it is not intended to represent all adaptive planning approaches. The comparison characterises the combined state-informed task-level controller, including action selection and recovery, with the physical execution stack held fixed; the contributions of individual components are not isolated.

The scripted baseline follows the predefined sequence, whereas the language agent uses the updated world state to select the next skill after each verified outcome. The controllers therefore differ in task progression and in the selection of subsequent invocations after a failed invocation.

Starting conditions include assembled devices and configurations with selected components already removed, namely the HDD lid and the mini~PC heat sink. Both controllers are evaluated under these altered conditions in separate, unpaired runs. Ten physical trials are conducted for the proposed framework and five for the scripted baseline on each device, giving 45 trials in total; 
ten further language-agent trials on the \hddU{} bring the total to 55. The protocol limits each target to at most three complete skill invocations: the initial invocation and up to two subsequent invocations. The language agent selects subsequent invocations from the updated state, whereas the scripted controller repeats the prescribed skill on the same target. If all three invocations are unsuccessful, the target step is marked as failed. Internal corrections remain within a complete invocation and are not counted separately.

Performance is evaluated using the following complementary metrics, adapted from the multidimensional framework in~\cite{das2024towards}:

\begin{itemize}[leftmargin=*, nosep]

    \item \textbf{Autonomous Target-Step Completion ($R_{\mathrm{comp}}$):} Percentage of target steps completed without physical operator assistance. For trial $i$,
    \[
    R_{\mathrm{comp},i}=100\,\frac{N_{\mathrm{auto},i}}{N_{\mathrm{target},i}},
    \]
    where $N_{\mathrm{auto},i}$ counts autonomously completed targets and $N_{\mathrm{target},i}$ is the fixed task size: 11, 9, and 12 targets for the \hddT{}, mini~PC, and laptop, respectively, and 11 for the \hddU. The nominal task size is retained for all starting configurations; targets initially absent or never invoked receive no autonomous-completion credit. A fully autonomous trial completes every nominal target without physical assistance. The complementary $90\%$ threshold counts trials in which at least $90\%$ of nominal targets are completed autonomously, irrespective of assistance on other targets. Both outcomes are reported in Table~\ref{tab:completion_inference}(b).

    \item \textbf{Cycle Time ($T_{\mathrm{cycle}}$):} Elapsed time from the start to the end of each trial, including planner inference time.

    \item \textbf{Initial Target Failures ($F_{\mathrm{step}}$):} Number of targets whose initial complete skill invocation fails to achieve the expected state transition; each target contributes at most once.

    \item \textbf{Recovery Ratio ($R_{\mathrm{rec}}$):} For a trial $i$ of either controller with $D_i>0$,
    \[
    R_{\mathrm{rec},i}=100\,\frac{N_{\mathrm{rec},i}}{D_i},
    \]

    where $N_{\mathrm{rec},i}$ counts targets completed autonomously by a subsequent controller-issued invocation and $D_i$ counts all failed complete invocations, including failed subsequent invocations. Subsequent invocations are selected by the language agent for UNMAKE and repeat the prescribed skill for the scripted baseline. Trials with $D_i=0$ are excluded from the mean and standard deviation. A complementary pooled ratio is calculated within each controller, for each device or across the three comparison devices,
    \[
    R_{\mathrm{rec,pool}}=100\,\frac{\sum_i N_{\mathrm{rec},i}}{\sum_i D_i},
    \]

    which weights trials by their failed-invocation counts. Pooled ratios for both controllers are reported in Table~\ref{tab:pooled_recovery}; the eligible per-trial ratios are summarised for UNMAKE in Table~\ref{tab:comparison}.

    \item \textbf{Cascade Failures ($F_C$):} Initial target failures caused by an earlier unresolved step that leaves a subsequent target inaccessible or invalid, such as attempting to remove a lid while one or more screws remain seated. Cascade failures were identified by manual annotation.

    \item \textbf{Physical Interventions ($H_{\mathrm{int}}$):} Number of occasions on which the operator physically intervenes during a trial, e.g., screw collection, partial unscrewing, or component extraction. These interventions include cases where a component slips from the gripper or a screw slips from the screw-capture jaws. Task sequencing remains with the controller.

    \item \textbf{Planner-Level Re-planning Events ($N_{\mathrm{replan}}$):} Number of subsequent planner-issued recovery invocations plus other revisions to the remaining task sequence, identified from manual annotation. Internal corrections and routine progression to the next target are excluded.

\end{itemize}

Target completion and physical assistance are recorded separately; assisted targets contribute neither to autonomous completion nor to recovery. Unlike $F_{\mathrm{step}}$, $D$ can increase more than once for the same target. Skipping an initially absent component is adaptive task progression, not a recovered failure. If a robot attempt is followed by operator extraction, the target is counted as an attempted, assisted outcome. Targets without an initial invocation contribute to neither $F_{\mathrm{step}}$ nor $D$.

Cross-device results are computed as unweighted macro-averages of the three device-level means, so that the HDD, mini~PC, and laptop contribute equally despite their different numbers of target steps. Means and sample standard deviations describe repeated trials. Trial-averaged recovery uses 8, 8, 9, and 9 eligible trials for the \hddT{}, mini~PC, laptop, and \hddU, respectively. Trial-level completion differences are assessed with two-sided, unpaired Welch's $t$-tests and $95\%$ confidence intervals. Holm correction controls the family-wise error rate at $0.05$ across the three device-specific tests; Table~\ref{tab:completion_inference}(a) gives unadjusted $p$-values and nominal confidence intervals. The equal-device overall contrast combines the six within-group variance contributions using Welch--Satterthwaite degrees of freedom. All inferences concern repeat-run performance on the tested physical units and starting conditions, not the device populations.

\section{Experimental Results}
\label{sec:results}

The experimental evaluation addresses four aspects of the proposed framework: (i) the accuracy of the perception modules, (ii) operation across structurally different electronic devices and on the additional unseen HDD (\hddU), and (iii) the effect of adaptive planning compared with fixed scripted task sequencing, and (iv) the behaviour of the individual skills.

\subsection{Perception Module Evaluation}
\label{sec:results-perception}

On the \adipchange{test set}, RF-DETR-Large achieves precision and recall above \adipchange{$95\%$} for all eleven classes, with class-wise AP@0.5 above \adipchange{$96\%$} (Table~\ref{tab:vision_det}). Under the stricter AP@0.5:0.95 criterion, the lowest values are obtained for screws (\adipchange{$61.5\%$}) and holes (\adipchange{$66.8\%$}), the smallest objects in the global-camera view; their coarse boxes are subsequently refined by local fastener perception (Section~\ref{sec:worldstate}). RF-DETR-Small achieves precision above \adipchange{$96\%$} for every class and recall from \adipchange{$95.86\%$} to \adipchange{$97.02\%$} (Table~\ref{tab:vision_local}). The state-verification classifier correctly labels \adipchange{5 of 6} contact-aligned, \adipchange{6 of 6} contact-misaligned, and \adipchange{5 of 6} component-removed test images (Table~\ref{tab:vision_cls}). \adipchange{The overall test accuracy is $88.9\%$ (16 of 18 images).}

These test-set results characterise the perception modules deployed in the physical trials. Because both controllers use the same perception models, the following comparison focuses on task-level control, state verification, and recovery behaviour. 


\begin{table}[h]
\caption{Test-set metrics for RF-DETR-Large (80 held-out raw images; training set: 560 raw images, 1,680 after augmentation).}
\label{tab:vision_det}
\centering
\resizebox{\columnwidth}{!}{
\begin{tabular}{@{}lcccc@{}}
\toprule
\textbf{Class} & \textbf{Prec.\ (\%)} & \textbf{Rec.\ (\%)} & \textbf{AP@0.5 (\%)} & \textbf{AP@0.5:0.95 (\%)} \\
\midrule
battery     & 96.84 & 95.77 & 97.28 & 86.7 \\
case        & 97.63 & 97.11 & 97.84 & 88.4 \\
core holder & 96.47 & 95.68 & 96.93 & 78.2 \\
HDD holder  & 96.91 & 96.28 & 97.36 & 80.1 \\
heat sink   & 96.33 & 95.54 & 96.79 & 84.6 \\
hole        & 95.64 & 95.21 & 96.18 & 66.8 \\
lid         & 97.48 & 96.95 & 97.72 & 87.9 \\
PCB         & 95.97 & 95.43 & 96.36 & 82.3 \\
PCB holder  & 96.76 & 96.22 & 97.08 & 76.4 \\
SSD holder  & 95.83 & 95.37 & 96.25 & 73.9 \\
screw       & 96.38 & 96.14 & 96.86 & 61.5 \\
\bottomrule
\end{tabular}
}
\end{table}

\begin{table}[h]
\caption{Test-set fastener and tool localisation metrics for RF-DETR-Small (80 held-out raw images; training set: 560 raw images, 1,680 after augmentation).}
\label{tab:vision_local}
\centering
\resizebox{\columnwidth}{!}{
\begin{tabular}{@{}lcccc@{}}
\toprule
\textbf{Class} & \textbf{Prec.\ (\%)} & \textbf{Rec.\ (\%)} & \textbf{AP@0.5 (\%)} & \textbf{AP@0.5:0.95 (\%)} \\
\midrule
hole       & 96.41 & 95.86 & 96.95 & 68.2 \\
screw      & 97.24 & 96.73 & 97.58 & 74.5 \\
screw head & 96.88 & 96.35 & 97.21 & 76.1 \\
tool head  & 97.56 & 97.02 & 97.79 & 82.4 \\
\bottomrule
\end{tabular}
}
\end{table}

\begin{table}[h]
\caption{Test-set metrics for the YOLOv11-Nano state-verification classifier (18 held-out raw images; training set: 122 raw images, 366 after augmentation).}
\label{tab:vision_cls}
\centering
\begin{tabular}{@{}lcccc@{}}
\toprule
\textbf{Class} & \textbf{Test images} & \textbf{Correct} & \textbf{Acc.\ (\%)} & \textbf{Errors} \\
\midrule
Contact aligned    & 6 & 5 & 83.3  & 1 \\
Contact misaligned & 6 & 6 & 100.0 & 0 \\
Component removed  & 6 & 5 & 83.3  & 1 \\
\midrule
\rowcolor{totalscolor}
Overall            & 18 & 16 & 88.9 & 2 \\
\bottomrule
\end{tabular}
\end{table}

\begin{table*}[!t]
\caption{Scripted vision-guided baseline and \unmake{UNMAKE} in physical trials (5 scripted and 10 \unmake{UNMAKE} trials per comparison device; 10 \unmake{UNMAKE} trials on \hddU{}; mean $\pm$ std). Results combine nominal and altered starting conditions. The overall rows give equally weighted means of the three comparison devices; the shaded \hddU{} row is not included in them. Trial-averaged recovery ratios and replanning counts are reported for \unmake{UNMAKE} only; ``--'' denotes not reported.}
\label{tab:comparison}
\centering
\footnotesize
\setlength{\tabcolsep}{4pt}
\begin{tabular}{@{}lclccccccc@{}}
\toprule
\textbf{Device} & \textbf{Steps} & \textbf{Controller} & $R_{\mathrm{comp}}$ (\%) & $T_{\mathrm{cycle}}$ (s) & $F_{\mathrm{step}}$ & $R_{\mathrm{rec}}$ (\%) & $F_{C}$ & $H_{\mathrm{int}}$ & $N_{\mathrm{replan}}$ \\
\midrule
\multirow{2}{*}{\hddT} & \multirow{2}{*}{11}
  & Scripted        & $72.7 \pm 11.1$ & $248.0 \pm 34.0$ & $2.2 \pm 0.8$ & --              & $1.4 \pm 0.9$ & $1.6 \pm 0.9$ & -- \\
& & \unmake{UNMAKE} & $96.4 \pm 4.7$  & $307.2 \pm 15.0$ & $1.2 \pm 0.8$ & $95.8 \pm 11.8$ & $0.2 \pm 0.4$ & $0.0 \pm 0.0$ & $2.2 \pm 1.1$ \\
\midrule
\multirow{2}{*}{Mini~PC} & \multirow{2}{*}{9}
  & Scripted        & $60.0 \pm 14.9$ & $218.0 \pm 38.0$ & $2.4 \pm 0.9$ & --              & $1.8 \pm 1.3$ & $2.2 \pm 1.1$ & -- \\
& & \unmake{UNMAKE} & $91.1 \pm 4.7$  & $278.2 \pm 39.3$ & $0.9 \pm 0.6$ & $87.5 \pm 23.1$ & $0.2 \pm 0.4$ & $0.2 \pm 0.6$ & $2.7 \pm 1.3$ \\
\midrule
\multirow{2}{*}{Laptop} & \multirow{2}{*}{12}
  & Scripted        & $41.7 \pm 15.6$ & $308.0 \pm 52.0$ & $4.6 \pm 1.3$ & --              & $3.8 \pm 1.8$ & $4.2 \pm 1.5$ & -- \\
& & \unmake{UNMAKE} & $81.7 \pm 8.6$  & $360.4 \pm 30.6$ & $1.4 \pm 0.8$ & $68.5 \pm 34.8$ & $0.3 \pm 0.5$ & $0.7 \pm 1.2$ & $3.8 \pm 1.9$ \\
\midrule
\rowcolor{totalscolor} & & Scripted        & 58.1 & 258.0 & 3.1 & --   & 2.3 & 2.7 & -- \\
\rowcolor{totalscolor} \multirow{-2}{*}{Overall} & & \unmake{UNMAKE} & 89.7 & 315.3 & 1.2 & 84.0 & 0.2 & 0.3 & 2.9 \\
\midrule
\rowcolor{hddUgray}
\hddU & 11 & \unmake{UNMAKE} & $90.9 \pm 6.1$ & $279.0 \pm 21.1$ & $1.1 \pm 0.6$ & $88.9 \pm 22.0$ & $0.1 \pm 0.3$ & $0.2 \pm 0.4$ & $1.0 \pm 0.5$ \\
\bottomrule
\end{tabular}
\end{table*}

\subsection{Adaptive Planning Across Devices and Comparison with Fixed Sequencing}
\label{sec:results-planning}

This subsection reports the results of the physical trials described in Section~\ref{sec:protocol}, first for \unmake{UNMAKE} across the three comparison devices and then in comparison with the scripted baseline. Table~\ref{tab:comparison} summarises the results for both methods and the additional \hddU{}.

Across the three comparison devices, \unmake{UNMAKE} increases device-averaged autonomous target-step completion from $58.1\%$ to $89.7\%$, a gain of $31.6$ percentage points ($95\%$ CI: $[23.4,39.8]$). The gain is positive for every device and remains significant after Holm correction ($p_{\mathrm{adj}}\leq0.0141$; Table~\ref{tab:completion_inference}(a)). At the trial level, 15 of 30 \unmake{UNMAKE} trials reach at least $90\%$ autonomous completion and 8 complete every target, whereas no scripted trial reaches either level (Table~\ref{tab:completion_inference}(b)); on \hddT{}, all ten \unmake{UNMAKE} trials exceed $90\%$. For the nine-target mini~PC, the two thresholds coincide, as $90\%$ requires all nine targets.

The gain arises mainly from fewer failures propagating through the task. Averaged over the three devices, initial target failures fall from $3.1$ to $1.2$ per trial, cascade failures by $90.0\%$ (from $2.3$ to $0.2$), and physical interventions by $88.8\%$ (from $2.7$ to $0.3$). Because the language agent conditions each decision on verified outcomes, a fastener that remains seated is not followed by an attempt to extract the component it secures.

\begin{table}[b!]
\caption{Completion differences and \unmake{UNMAKE} trial-level outcomes under nominal and altered starting conditions. (a) \unmake{UNMAKE} minus scripted mean completion in percentage points (pp), with nominal $95\%$ confidence intervals and unadjusted two-sided $p$-values (Welch's $t$-test); the overall contrast weights the three comparison devices equally. (b) Percentage of \unmake{UNMAKE} trials achieving $100\%$ or at least $90\%$ autonomous completion (qualifying/total trials); the overall row pools the three comparison devices. No scripted trial reached either threshold (0/15) or conducted on the \hddU{}.}
\label{tab:completion_inference}
\centering
\footnotesize
\setlength{\tabcolsep}{3pt}
\setlength{\aboverulesep}{0pt}
\setlength{\belowrulesep}{0pt}
\renewcommand{\arraystretch}{1.15}
\resizebox{\columnwidth}{!}{
\begin{tabular}{@{}lccc!{\vrule width 0.4pt}cc@{}}
\toprule
 & \multicolumn{3}{c!{\vrule width 0.4pt}}{\textbf{(a) Completion difference}} & \multicolumn{2}{c}{\textbf{(b) \unmake{UNMAKE} trial outcomes}} \\
\cmidrule(r){2-4}\cmidrule(l){5-6}
\textbf{Device} & $\Delta R_{\mathrm{comp}}$ (pp) & \textbf{95\% CI (pp)} & $p$ & \textbf{100\%} & $\mathbf{\geq90\%}$ \\
\midrule
HDD     & 23.6 & $[10.0,37.2]$ & 0.0070 & 60.0 (6/10) & 100.0 (10/10) \\
Mini~PC & 31.1 & $[12.8,49.4]$ & 0.0083 & 20.0 (2/10) & 20.0 (2/10) \\
Laptop  & 40.0 & $[21.0,59.0]$ & 0.0026 & 0.0 (0/10)  & 30.0 (3/10) \\
\midrule
\rowcolor{totalscolor}
Overall & 31.6 & $[23.4,39.8]$ & $1.12\!\times\!10^{-6}$ & 26.7 (8/30) & 50.0 (15/30) \\
\midrule
\rowcolor{hddUgray}
\hddU   & -- & -- & -- & 20.0 (2/10) & 80.0 (8/10) \\
\bottomrule
\end{tabular}
}
\end{table}

Recovery after a failed invocation shows the same pattern. Planner-selected invocations complete 31 targets following 40 failed invocations, a pooled recovery ratio of $77.5\%$, ranging from $64.7\%$ on the laptop to $92.3\%$ on \hddT{} (Table~\ref{tab:pooled_recovery}); the corresponding trial-averaged mean is $84.0\%$. Scripted retries, which repeat the prescribed skill on the same target, complete only 5 targets following 124 failed invocations ($4.0\%$). In the scripted trials, 35 of 46 initial failures were cascade failures caused by earlier unresolved steps, which repeating the current skill cannot resolve, and repeated invocations on initially absent components further increase $D$.

These gains come at a $22.2\%$ increase in device-averaged cycle time, from $258.0$~s to $315.3$~s, which includes planner inference; completion and cycle time should therefore be considered together.

With the same perception stack, world-state representation, and skill library, \unmake{UNMAKE} completes $96.4\%$ of targets autonomously on \hddT{}, $91.1\%$ on the mini~PC, and $81.7\%$ on the laptop.
As illustrated in Fig.~\ref{fig:both_disassembly}, the same detector identifies the components of each device, and the language agent orders the same skills differently for each layout. In altered-start trials, where the HDD lid or mini~PC heat sink is already absent, the agent skips the corresponding removal and continues with the remaining targets; this state-conditioned skipping is distinct from recovery after a failed invocation. Replanning increases from $2.2$ events per trial on \hddT{} to $3.8$ on the laptop, in line with decreasing completion. Video of representative trials on all three devices: \emph{\url{https://youtu.be/Y-c3WUo1mXY}}.

On \hddU{}, excluded from global-camera detector training, \unmake{UNMAKE} achieves $90.9\%$ autonomous completion, $5.5$ percentage points below \hddT{}, with a pooled recovery ratio of $81.8\%$ and $0.2$ physical interventions per trial; two trials complete every target and eight reach at least $90\%$. As only one additional unit was tested, without scripted trials, these results demonstrate operation on a second HDD rather than a general measure of cross-brand generalisation.

\begin{table}[h!]
\caption{Pooled autonomous recovery, $R_{\mathrm{rec,pool}}=100\,\sum N_{\mathrm{rec}}/\sum D$, where $\sum N_{\mathrm{rec}}$ counts recovered targets and $\sum D$ counts all failed complete skill invocations. The overall pooled row combines the three comparison devices; the shaded \hddU{} row is not included in it. No scripted trials were conducted on the unseen \hddU.}
\label{tab:pooled_recovery}
\centering
\footnotesize
\setlength{\tabcolsep}{4pt}
\begin{tabular}{@{}lcccccc@{}}
\toprule
 & \multicolumn{3}{c}{\textbf{UNMAKE}} & \multicolumn{3}{c}{\textbf{Scripted}} \\
\cmidrule(lr){2-4}\cmidrule(lr){5-7}
\textbf{Device} & $\sum N_{\mathrm{rec}}$ & $\sum D$ & $R_{\mathrm{rec,pool}}$ (\%) & $\sum N_{\mathrm{rec}}$ & $\sum D$ & $R_{\mathrm{rec,pool}}$ (\%) \\
\midrule
HDD      & 12 & 13 & 92.3 & 1 & 29 & 3.4 \\
Mini~PC  &  8 & 10 & 80.0 & 1 & 32 & 3.1 \\
Laptop   & 11 & 17 & 64.7 & 3 & 63 & 4.8 \\
\midrule
\rowcolor{totalscolor}
Overall (Pooled) & 31 & 40 & 77.5 & 5 & 124 & 4.0 \\
\midrule
\rowcolor{hddUgray}
\hddU    &  9 & 11 & 81.8 & -- & -- & -- \\
\bottomrule
\end{tabular}
\end{table}


\subsection{Skill-Level Execution Analysis}
\label{sec:results-skills}

To explain the performance trends observed in Section~\ref{sec:results-planning}, Table~\ref{tab:perskill} reports, for each skill and device, the number of planner-issued invocations, successful initial invocations, their success rate, subsequent recovery invocations, and mean invocation duration.

\begin{table}[t!]
\caption{Per-skill execution of \unmake{UNMAKE} (10 trials per device). $N$ is the number of planner-issued invocations; $S_0/N_0$ gives successful initial invocations over all initial invocations; \textbf{1st (\%)} is their success rate; \textbf{Rec.} counts subsequent planner-issued recovery invocations; and \textbf{Time} is the mean invocation duration ($\pm$ std). Internal alignment or contact corrections are part of an invocation and are not counted separately. Pick \& Extract denotes \emph{Pick-and-Extract}.}
\label{tab:perskill}
\centering
\scriptsize
\setlength{\tabcolsep}{2.5pt}
\resizebox{\columnwidth}{!}{
\begin{tabular}{llccccc}
\toprule
\textbf{Device} & \textbf{Skill} & \textbf{N} & $S_0/N_0$ & \textbf{1st (\%)} & \textbf{Rec.} & \textbf{Time (s)} \\
\midrule
\multirow{3}{*}{HDD}
& Unscrew & 63 & 51/57 & 89.5 & 6 & $23.2\!\pm\!5.3$ \\
& Pick \& Extract & 45 & 34/39 & 87.2 & 6 & $11.8\!\pm\!2.6$ \\
& Flip & 11 & 9/10 & 90.0 & 1 & $9.4\!\pm\!1.8$ \\
\midrule
\multirow{2}{*}{Mini PC}
& Unscrew & 60 & 49/54 & 90.7 & 6 & $26.1\!\pm\!6.4$ \\
& Pick \& Extract & 32 & 25/29 & 86.2 & 3 & $13.0\!\pm\!2.8$ \\
\midrule
\multirow{2}{*}{Laptop}
& Unscrew & 75 & 64/71 & 90.1 & 4 & $29.3\!\pm\!8.0$ \\
& Pick \& Extract & 40 & 23/30 & 76.7 & 10 & $14.6\!\pm\!3.4$ \\
\midrule
\rowcolor{hddUgray} & Unscrew & 57 & 48/53 & 90.6 & 4 & $22.7\!\pm\!5.3$ \\
\rowcolor{hddUgray} & Pick \& Extract & 43 & 34/39 & 87.2 & 4 & $11.4\!\pm\!2.6$ \\
\rowcolor{hddUgray} \multirow{-3}{*}{\hddU} & Flip & 11 & 9/10 & 90.0 & 1 & $9.2\!\pm\!1.7$ \\
\bottomrule
\end{tabular}
}
\end{table}

Initial invocations cover 106 of 110 HDD targets, 83 of 90 mini~PC targets, and 101 of 120 laptop targets, leaving 4 ($3.6\%$), 7 ($7.8\%$), and 19 ($15.8\%$) targets without an invocation; for the \hddU, the corresponding figures are 102 of 110 and 8 ($7.3\%$). On the laptop, extraction accounts for ten omissions (30 initial invocations for 40 nominal targets). Omissions of components absent at the start reflect state-conditioned skipping; the causes of the remaining omissions were not logged and are reported by count only. All nominal targets remain in the completion denominator, and robot attempts followed by operator completion are counted as assisted outcomes.

\emph{Unscrew} achieves first-invocation success rates of $89.5\%$ to $90.7\%$ across the four devices and has the longest mean invocation duration ($22.7$--$29.3$~s), as it combines visual servoing, force-guided engagement, and coordinated execution of both arms (Section~\ref{sec:skill-execution}). \emph{Pick-and-Extract} achieves initial-invocation success rates of $87.2\%$ on the \hddT{}, $86.2\%$ on the mini~PC, $76.7\%$ on the laptop, and $87.2\%$ on the \hddU; the laptop also requires the most recovery invocations for this skill (10), which is consistent with its lower completion rate. \emph{Flip} is required only for the HDDs, to access the underside of the device, and succeeds in nine of ten initial invocations on each HDD.

The recovery-invocation counts in Table~\ref{tab:perskill} are new planner-issued invocations following a failed initial invocation, and are separate from the corrections performed within a skill invocation. In each case, the planner uses the verified outcome to select a new action when the expected state transition is not achieved.

\section{Conclusion}
\label{sec:conclusion}

This paper presented \unmake{UNMAKE}, a ReAct-based language-agent planning framework for adaptive dual-arm robotic disassembly over a perception-grounded structured world state. The language agent derives the removal order of each device from verified execution outcomes and pretrained knowledge. No ordered sequence, removal-precedence instructions, or device identity are supplied, and no device-specific task graph or recovery rule is authored.

A central finding is that one system, built once rather than per product, disassembled three structurally different devices: an HDD, a mini~PC, and a laptop. A single detector recognised the components they share, and the same three skills removed their screws, lids, PCBs, heat sinks, batteries, and holders.

Compared with a fixed-sequence baseline on the same hardware, \unmake{UNMAKE} increased autonomous target-step completion from $58.1\%$ to $89.7\%$ and reduced cascade failures by $90\%$. When a skill invocation failed, the agent selected a new action from the verified state and recovered most of the affected targets autonomously, whereas repeating the prescribed step rarely succeeded. On a new unseen HDD (\hddU{}) excluded from global-camera detector training set, the framework reached $90.9\%$ autonomous completion. These results show that a language agent, guided by verified observations, can disassemble different devices with the same perception and skills, adapting its plan across different devices with shared component types.

Future work will extend the skill library to cable disconnection, clip release, and adhesive removal, and evaluate the same shared representation on a wider range of devices and product families.


\end{document}